%% file: iclr2027_conference.tex
\documentclass{article} 
\usepackage{iclr2027_conference,times}

\input{math_commands.tex}

\usepackage{hyperref}
\usepackage{url}

\usepackage[utf8]{inputenc} 
\usepackage[T1]{fontenc}    
\usepackage{booktabs}       
\usepackage{amsfonts}       
\usepackage{nicefrac}       
\usepackage{microtype}      
\usepackage{xcolor}         
\usepackage{graphicx}
\usepackage{booktabs}
\usepackage{colortbl}
\usepackage{multirow}
\usepackage{array}
\usepackage{wrapfig}
\usepackage{amsmath}
\usepackage{enumitem}
\usepackage{bbm}
\usepackage{mwe}
\usepackage[most]{tcolorbox}
\usepackage{amsmath,amssymb,bm,booktabs}
\usepackage{cleveref}

\title{BATCH: \underline{B}ehavior \underline{A}lignmen\underline{t} for \underline{C}ross-Embodiment Learning from \underline{H}umans}

\author{Zeyi Liu ~~ Shuran Song \\[2mm]
    Stanford University \\[2mm]
    \textup{\url{https://batch-embodiment.github.io}}
}

\iclrfinalcopy 
\begin{document}

\maketitle

\vspace{-2mm}
\begin{abstract}
Human demonstrations provide a scalable source of diverse manipulation data, but leveraging them for robot learning remains challenging due to differences in embodiment, viewpoint, action spaces, and temporal structure. We introduce \textbf{BATCH}, a cross-embodiment learning framework that explicitly aligns human and robot behaviors. BATCH learns self-supervised temporal representations with a cycle-consistency objective to establish cross-embodiment correspondences without manual annotation. We further introduce two complementary training objectives that leverage these correspondences within an autoregressive video-action diffusion model: behavior-aligned co-training and human-to-robot style transfer. On real-world manipulation tasks, BATCH improves generalization to unseen task configurations compared with training on unaligned data and prior cross-embodiment baselines, with additional gains from the style-transfer objective. Our results demonstrate that explicit behavior alignment enables more effective transfer from human demonstrations to robot policies.
\end{abstract}

\setlength{\parskip}{4pt}
\input{text/intro_v2}
\input{text/related_work}
\input{text/method}
\input{text/eval}

\textbf{Limitation. } Our alignment approach assumes shared behavioral structure between human and robot demonstrations and may struggle when strategies or action orderings differ substantially. Combining VLM-based coarse semantic alignment with our frame-level refinement could help address this limitation. Alignment errors may also introduce mismatched supervision into the style-transfer objective. Finally, BATCH requires robot demonstrations during post-training with human data; reducing this dependence remains an important direction for future work.

\vspace{-1mm}
\section{Conclusion}
\vspace{-1mm}
We present \textbf{BATCH}, a cross-embodiment learning framework that combines \emph{behavior alignment} with a \emph{cross-embodiment video-action model}. We use GTCC with DTW to temporally align human and robot behaviors in a learned embedding space, then train on the aligned demonstrations with joint co-training and human-to-robot style transfer. Across two real-world manipulation tasks, training with aligned data consistently outperforms unaligned training, while style transfer provides further gains beyond co-training alone. These results demonstrate the value of explicitly establishing and leveraging behavior correspondences for effective human-to-robot transfer.
\clearpage

\setlength{\parskip}{.5pc}
\input{text/addn}
\bibliography{references}
\bibliographystyle{iclr2027_conference}

\newpage
\appendix
\input{text/supp}

\end{document}

%% file: math_commands.tex
\usepackage{amsmath,amsfonts,bm}

\def\eqref#1{equation~\ref{#1}}

\def\1{\bm{1}}

\DeclareMathAlphabet{\mathsfit}{\encodingdefault}{\sfdefault}{m}{sl}
\SetMathAlphabet{\mathsfit}{bold}{\encodingdefault}{\sfdefault}{bx}{n}

\newcommand{\KL}{D_{\mathrm{KL}}}



%% file: text/intro_v2.tex
\vspace{-2mm}
\section{Introduction}
\vspace{-2mm}
\begin{figure}[h]
    \centering
    \includegraphics[width=\linewidth]{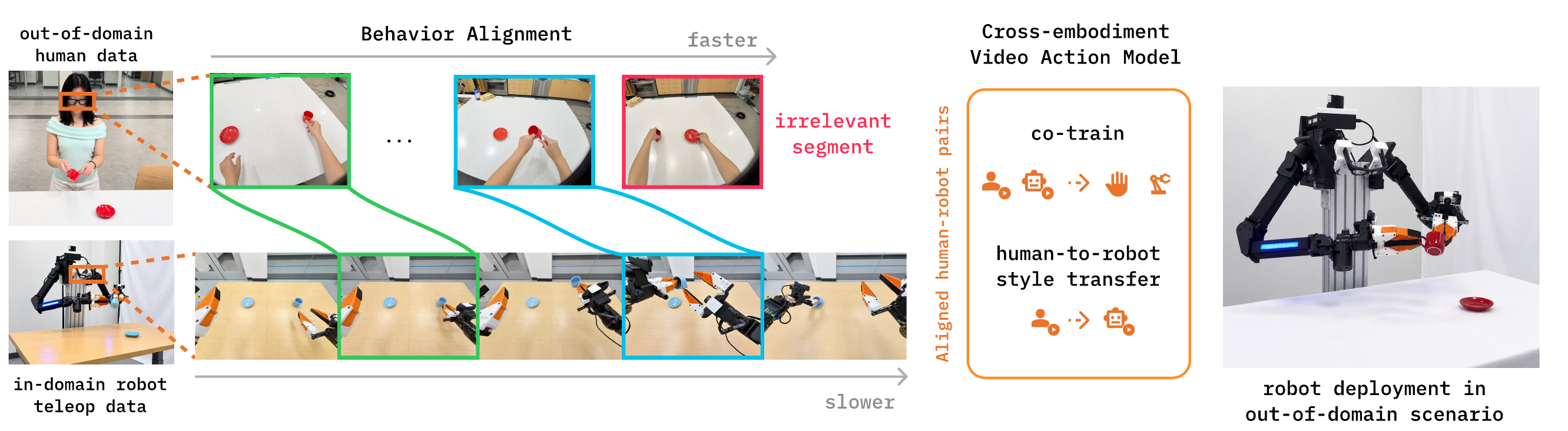}
    \vspace{-6mm}
    \caption{\small \textbf{BATCH} performs behavior alignment between diverse out-of-domain human data and limited in-domain robot teleop data, accounting for differences in execution speed and irrelevant human segments. The aligned pairs are leveraged through two objectives within a downstream cross-embodiment video action model: co-training and human-to-robot style transfer. The resulting policy transfers visual knowledge from human data to robot deployment in out-of-domain scenarios.
    }
    \vspace{-2mm}
    \label{fig:teaser}
\end{figure}

\newtcolorbox{questionbox}{
    colback=gray!8,
    colframe=gray!45,
    boxrule=0.5pt,
    arc=2pt,
    left=6pt,
    right=6pt,
    top=5pt,
    bottom=5pt,
    title=\textbf{Questions},
    coltitle=black,
    fonttitle=\small,
}

Human demonstrations offer a scalable way to expand the scene and behavior diversity of robot datasets. In particular, wearable egocentric sensing devices, such as smart glasses~\citep{engel2023project}, provide a lightweight and portable solution for capturing natural human manipulation across diverse environments, while on-device SLAM cameras enable accurate hand-pose tracking. Despite this promise, directly leveraging human demonstrations for robot learning remains challenging due to the substantial embodiment gap between humans and robots, including both visual appearances and physical motions.

Prior studies have shown that directly co-training on cross-embodiment data can improve visual robustness and policy generalization to unseen backgrounds and objects~\citep{kareer2025egomimic, yang2026data}. Yet cross-embodiment data is not simply additional data from a different visual domain: even when performing the same task, humans and robots can differ substantially in execution speed, temporal structure, action spaces, and visual observations due to differences in embodiment. Human demonstrations may also contain irrelevant segments that have no corresponding robot behavior. As a result, naively scaling human demonstrations does not necessarily provide useful supervision for robot behavior. Instead, models can separate the two embodiments and rely on embodiment-specific cues, limiting the extent to which human experience transfers to robot control. Empirically, this can lead to diminishing gains or even negative transfer as more poorly matched human data is incorporated~\citep{wei2025empirical, punamiya2026egoverse}. We therefore hypothesize that effective cross-embodiment transfer requires explicitly establishing behavioral correspondences between human and robot trajectories, providing a learning signal for identifying task-relevant structure that should transfer while discounting embodiment-specific variation.

We define \textbf{behavior alignment} as establishing temporal frame-level correspondences between equivalent phases of task execution across embodiments, based on task-relevant states and interactions rather than embodiment-specific appearance, viewpoint, or action representation. The key challenge is to discover these correspondences automatically, without requiring manually annotated phase labels or frame-level matches between human and robot trajectories. Achieving this requires features that capture behavioral progression over time while suppressing embodiment-specific variation. Prior non-learning approaches commonly apply dynamic time warping (DTW) over actions or static visual features, which can be unreliable under large embodiment gaps or when visual states are ambiguous. Instead, we learn self-supervised features with a {temporal cycle-consistency} objective~\citep{dwibedi2019temporal, donahue2024gtcc}, where each frame representation incorporates information from multiple neighboring frames to capture behavioral progression rather than relying on a single observation. We then apply subsequence DTW in this learned feature space to retrieve temporally corresponding human and robot segments, accommodating differences in execution speed and unmatched segments.

\looseness-1
Another key contribution of our approach is how we leverage the aligned trajectories for cross-embodiment learning. We instantiate our method within an autoregressive video-action diffusion model~\citep{ye2026world} pretrained on in domain robot demonstrations, and finetune with two complementary objectives over aligned human and robot trajectories. A behavior-aligned co-training objective learns visual dynamics and motions from both embodiments, while a cross-embodiment style-transfer objective translates human observations into temporally corresponding robot-domain videos. Together, these objectives learn from both embodiments while explicitly coupling them under matched behavior, enabling more effective transfer from diverse human demonstrations to robot policies.

We evaluate our approach on two manipulation tasks from EgoVerse~\citep{punamiya2026egoverse}: single-arm \textit{Object in Container}, which requires continuous execution, and bimanual \textit{Cup on Saucer}, which requires precise grasping and bimanual coordination. We consider out-of-domain test scenarios presented in the human demonstrations but absent from the robot training data. Across the two tasks, our approach improves out-of-domain policy performance by an average of 45\% over co-training alone and 138\% over the robot-only pretrained model. Removing behavior alignment substantially degrades performance, highlighting the importance of reliable human-robot correspondences, while the style-transfer objective provides further gains with aligned pairs. These results demonstrate that both establishing behavioral correspondences and effectively leveraging them are critical for cross-embodiment transfer.

%% file: text/related_work.tex
\vspace{-2mm}
\section{Related Work}
\vspace{-2mm}
\label{related_work}

\textbf{Video Temporal Alignment. }
Video temporal alignment establishes frame-level correspondences across videos of similar behaviors despite variations in embodiment, viewpoint, and appearance. One foundational class of approaches is Temporal Cycle-Consistency (TCC)~\citep{dwibedi2019temporal,hadji2021representation}, which learns a shared embedding space from unaligned videos without manual labels by enforcing cycle-consistent nearest-neighbor matches. GTCC~\citep{donahue2024gtcc} further improves the robustness of TCC through multi-neighbor, multi-cycle matching modeled with Gaussian mixtures. In robotics, TCC-style representations have been used to derive dense task-progress rewards for cross-embodiment learning~\citep{zakka2022xirl}. Separately, Dynamic Time Warping (DTW) is a common technique used to align cross-embodiment demonstrations with different execution speeds~\citep{punamiya2025egobridge,liu2025immimic,yang2026data}. Our approach combines GTCC and DTW: GTCC learns cross-video correspondences in a shared embedding space, while DTW produces globally consistent temporal alignments across human and robot demonstrations while accounting for different execution speeds.


\textbf{Cross-Embodiment Learning from Human Data. }
Human demonstrations provide scalable manipulation data with greater scene and behavior diversity than costly robot demonstrations, but effective transfer remains challenging due to embodiment gap. Prior attempts include learning shared skill representations~\citep{xu2023xskill,pmlr-v305-kim25d}, using transferable intermediate representations such as object flow~\citep{pmlr-v270-xu25a}, or editing human videos into robotized demonstrations~\citep{lepert2025masquerade, lepert2025phantom, wang2026ego2robot}. However, these approaches can be limited in scalability with additional processing and tuning. More recent approaches directly co-train on human and robot data~\citep{kareer2025egomimic,qiu2025humanoid,yuan2025motiontrans,tao2025dexwild,zhu2026emma,punamiya2026egoverse}, with follow-up works seeking to improve cross-embodiment transfer through observation alignment~\citep{punamiya2025egobridge}, embodiment-agnostic visual prediction~\citep{li2026egowam}, or image- and action-space alignment~\citep{li2026matters}. Our work studies whether temporal behavior alignment between human and robot trajectories can further support cross-embodiment transfer while integrating naturally into existing co-training formulation.

\textbf{Video Action Models} jointly model future visual observations and robot actions, providing richer supervision by learning scene dynamics alongside robot control. Early methods such as GR-1 and GR-2 combine visual generation with action prediction~\citep{wu2024unleashing,cheang2024gr}. More recent approaches, including UVA~\citep{li2025uva}, UWM~\citep{zhu2025uwm}, and VideoVLA~\citep{shen2025videovla}, more tightly couple video and action modeling, while Cosmos Policy~\citep{kim2026cosmospolicy}, DreamZero~\citep{ye2026world}, LingBot-VA~\citep{li2026lingbot}, and LingBot-VA 2.0~\citep{zhang2026lingbotva2} scale this paradigm with large video backbones and video-action pretraining. Our work builds on this paradigm to study cross-embodiment policy learning from behaviorally aligned human and robot trajectories.

%% file: text/method.tex
\vspace{-1mm}
\section{Method}
\vspace{-2mm}
\label{sec:method}

\begin{figure}[htbp]
  \centering
  \includegraphics[width=\textwidth]{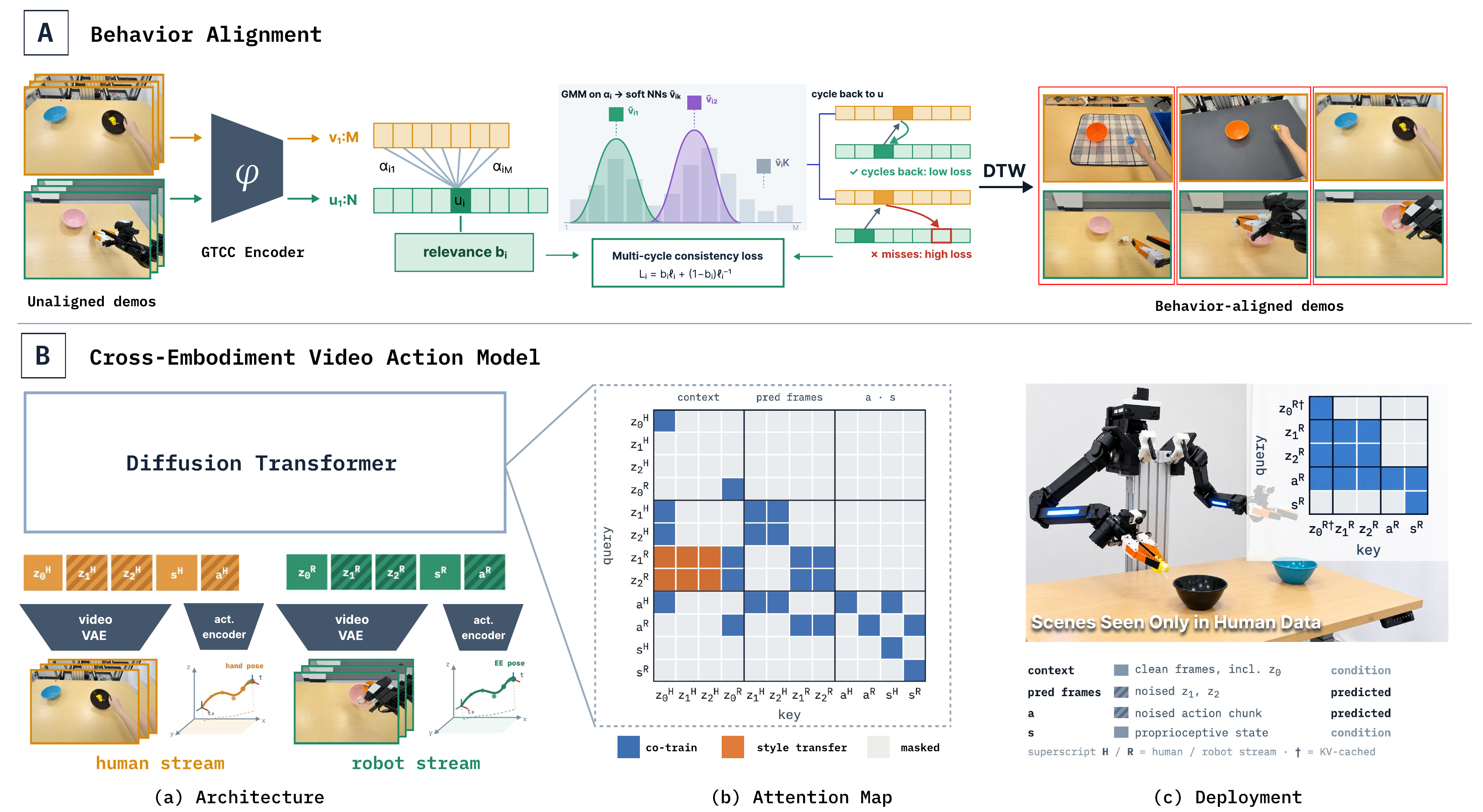}
\caption{
    \textbf{BATCH method.}
    (A) \emph{Behavior Alignment (\Cref{subsec:behavior-alignment})}: GTCC encoder trained with multi-cycle consistency learns shared temporal embeddings from unaligned demonstrations; the trained encoder then combines with DTW to retrieve behavior-aligned human and robot segments.
    (B) \emph{Cross-Embodiment Video Action Model} (\Cref{subsec:video-action-model}): (a) aligned human and robot trajectories are encoded as separate streams and input to a diffusion transformer; (b) the attention pattern supports both joint co-training and human-to-robot style transfer objective within the same model; and (c) at deployment, the robot stream is used to execute policy in scenarios only seen in human data.
}
  \vspace{-4mm}
  \label{fig:method}
\end{figure}


\subsection{Cross-Embodiment Behavior Alignment}
\label{subsec:behavior-alignment}

\looseness-1
We consider human and robot demonstrations of the same manipulation task. Following EgoVerse~\citep{punamiya2026egoverse}, human demonstrations are collected using Aria glasses, while robot demonstrations are collected through teleoperation (see Appendix~\Cref{app:setup} for data collection details). Despite sharing the same task objective, the two datasets can differ substantially in embodiment, background, viewpoint, and execution speed, making direct frame-level matching unreliable. We therefore seek a representation that is robust to these cross-embodiment differences while preserving temporal task progress. Starting from pre-extracted image features, we learn a shared temporal embedding space using the self-supervised cycle-consistency objective of \citet{donahue2024gtcc}, encouraging human and robot frames at corresponding stages of task execution to be close in the learned space. These correspondences are characterized by similar object states and interactions rather than visual appearance alone. To further account for differences in execution speed, we recover explicit frame-level correspondences using subsequence dynamic time warping (DTW), which finds the minimum-cost alignment in the learned embedding space while allowing non-uniform temporal warping.

\subsubsection{Problem Setup}
\label{sec:setup}

Given a set of robot demonstration segments \(\mathcal{D}^{\mathrm{r}}=\{X_i^{\mathrm{r}}\}_{i=1}^{N_{\mathrm{r}}}\) and human demonstration segments \(\mathcal{D}^{\mathrm{h}}=\{X_j^{\mathrm{h}}\}_{j=1}^{N_{\mathrm{h}}}\), where each segment corresponds to a temporal portion of the same manipulation task, our goal is to learn a shared encoder \(f_\theta\) that maps egocentric head-camera videos from both embodiments into sequences of per-frame embeddings. For a robot query segment \(X^{\mathrm{r}}\) of length \(N\) and a human target segment \(X^{\mathrm{h}}\) of length \(M\), we denote the corresponding embeddings as \(\mathbf{u}_{1:N}=f_\theta(X^{\mathrm{r}})\) and \(\mathbf{v}_{1:M}=f_\theta(X^{\mathrm{h}})\), respectively.

\textbf{Feature extraction. } We extract per-frame feature maps of shape \(1024\times14\times14\) using a frozen ImageNet ResNet-50 encoder \citep{he2016deep}. For each frame \(t\), features from the causal temporal window \([t-k+1,t]\) of length \(k\) are stacked and processed by two sequential 3D convolutional layers. The resulting representation is max-pooled over space and time and projected by a three-layer MLP to a 128-dimensional embedding.

\textbf{Multimodal soft nearest neighbors.}
As illustrated in \Cref{fig:method}, for each robot query frame \(i\), we compute a soft nearest-neighbor distribution over the human target sequence based on distance in the learned embedding space,
$$
\alpha_{ij}
=
\frac{\exp\!\left(-\lVert \mathbf{u}_i-\mathbf{v}_j\rVert_2^2/\tau\right)}
{\sum_{j'}\exp\!\left(-\lVert \mathbf{u}_i-\mathbf{v}_{j'}\rVert_2^2/\tau\right)},
\qquad j=1,\ldots,M,
$$
\(\tau\) denotes temperature. Standard TCC \citep{dwibedi2019temporal} summarizes this distribution with a single soft nearest neighbor, \(\tilde{\mathbf{v}}_i=\sum_j\alpha_{ij}\mathbf{v}_j\). However, manipulation trajectories may revisit similar task states, so a query frame can have multiple plausible correspondences at different target times. Averaging across these modes can therefore produce a representation that matches none of them. Following \citet{donahue2024gtcc}, we instead approximate \(\alpha_i\) with a \(K\)-component Gaussian mixture over target time,
$$
\min_{\{\pi_{ik},\mu_{ik},\sigma_{ik}\}_{k=1}^{K}}
\KL\!\left(
\alpha_i
\,\Big\|\,
\sum_{k=1}^{K}
\pi_{ik}\mathcal{N}(\cdot;\mu_{ik},\sigma_{ik}^2)
\right),
$$
where \(K\) is the number of mixture components, and \(\pi_{ik}\), \(\mu_{ik}\), and \(\sigma_{ik}\) denote the weight, temporal center, and spread of component \(k\). Each component defines a normalized distribution \(\hat{g}_{ik}\) over target frames and yields a separate soft nearest neighbor, \(\tilde{\mathbf{v}}_{ik}=\sum_j\hat{g}_{ikj}\mathbf{v}_j\). This preserves multiple plausible correspondence hypotheses rather than collapsing them into a single averaged match.

\textbf{Training objective. }
For each query frame \(i\), we minimize a multi-cycle consistency loss
$$
\ell_i
=
\sum_{k=1}^{K}
\pi_{ik}
\left[
(i-\hat{\mu}_{ik})^2
+
\lambda\log\hat{\sigma}_{ik}
\right],
$$
where \(\hat{\mu}_{ik}\) and \(\hat{\sigma}_{ik}\) denote the mean and uncertainty of the cycle-back distribution for correspondence hypothesis \(k\), and \(\lambda\) controls the uncertainty term. This objective encourages each hypothesis to cycle back to the original temporal location with low uncertainty. To handle frames without a meaningful counterpart in the other embodiment, an auxiliary head predicts a relevance weight \(b_i\) for each embedding, and we optimize \(L_i=b_i\ell_i+(1-b_i)\ell_i^{-1}\). Thus, alignable frames are encouraged to achieve low cycle-consistency error, while unalignable frames are down-weighted.


\subsubsection{Dataset Generation with DTW Retrieval}
\label{method:dataset}

To construct behavior-aligned human-robot pairs, we use subsequence DTW in the learned embedding space. Given a robot query window \(\mathbf{u}_{1:L}\) and a candidate human sequence \(\mathbf{v}_{1:T}\), we define the pairwise embedding cost as $C_{ij}=\|\mathbf{u}_i-\mathbf{v}_j\|_2$. Subsequence DTW finds the minimum-cost human sub-segment while allowing temporal warping to account for differences in execution speed. For each human candidate, we select the endpoint \(j^\star=\arg\min_j D[L,j]\), where \(D[L,j]\) is the cumulative DTW cost of aligning the full robot query to a human sub-segment ending at \(j\), and backtrack from \(j^\star\) to recover the matched sub-segment. We rank all human candidates by their normalized DTW cost and retain the top-\(k\) lowest-cost matches.

\vspace{-2mm}
\subsection{Cross-Embodiment Video Action Model}
\vspace{-2mm}
\label{subsec:video-action-model}

Given the aligned human and robot trajectories, we train a cross-embodiment video action model aiming to transfer task-relevant visual knowledge from human demonstrations to the robot domain. Our model (shown in \Cref{fig:method}B) builds on a video action diffusion transformer \citep{ye2026world} that jointly predicts future visual observations and chunks of robot actions. Training consists of two stages: we first pretrain on robot demonstrations, then finetune on aligned human-robot trajectories with two complementary objectives: (1) \emph{joint co-training}, which processes human and robot trajectories as attention-isolated streams with shared model parameters, and (2) \emph{human-to-robot style transfer}, which predicts robot future video from the aligned human video, explicitly encouraging human-to-robot visual transfer. At deployment, the model predicts actions from unseen robot observations observed only in human data.

\textbf{Backbone. }
Our policy builds on Wan2.1-I2V-14B \citep{wan2025wan}. We retain its first 8 transformer blocks, video VAE, and CLIP ViT-H/14 image encoder \citep{ilharco_gabriel_2021_5143773}. Since we consider a single-task setting, we remove text conditioning.

\textbf{Tokenization. } Each training sample consists of 9 head-camera frames, a 16-step future action chunk, the current proprioceptive state, and one conditioning frame from each wrist camera. For both human and robot demonstrations, states and actions are expressed in the coordinate frame of the Aria glasses. The video VAE compresses the 9 head-camera frames into three latent frames, where the first latent frame provides conditioning context and the remaining two form a causal block to be predicted.

For each embodiment, the transformer input is organized into four token groups: (1) \emph{clean video tokens}, obtained by patchifying uncorrupted latent frames on a shared canvas by concatenating the head and wrist latents along the width dimension; (2) \emph{noisy video tokens}, obtained by patchifying the noised head-camera latents; (3) a \emph{register} consisting of 16 action tokens and one state token, projected using embodiment-specific encoders; and (4) \emph{CLIP context tokens}, obtained by encoding the first frame from each available camera into 257 tokens per image and injected through cross-attention. The transformer predicts flow velocities at the noisy video and action token positions, with the corresponding prediction targets determined by the training objective. Token counts and remaining details are given in Appendix \Cref{app:tasks}.

\textbf{Human and robot streams. }
For each aligned human-robot pair sampled from the dataset described in \Cref{method:dataset}, we construct human and robot token streams and concatenate them into a single transformer sequence. Both streams share the same transformer weights; they differ only in their inputs and embodiment-specific state/action encoders and decoders. The robot stream receives head and wrist camera observations, whereas the human stream contains only the head-camera observation, with the missing wrist-camera regions zero-padded.

As shown in \Cref{fig:method} (b), attention within each stream follows a block-causal structure. Noisy video tokens attend to the clean visual context and to tokens within their own noisy video block. Action tokens additionally attend to the noisy video block and the current state token. We further design the two streams so that each can be processed independently of whether the other is present. First, rotary positional indices are reset at the beginning of each stream, preventing the positional representation of one embodiment from depending on the length of the other. Second, CLIP cross-attention is separated by embodiment: robot queries attend only to robot CLIP tokens, while human queries attend only to human CLIP tokens. As a result, concatenating the two streams does not alter the within-stream computation, while still allowing them to share weights.

\subsubsection{Training Objectives}
\label{sec:objectives}

We train video and action prediction with flow matching. A clean latent \(\mathbf{x}_0\) is corrupted to \(\mathbf{x}_\sigma=(1-\sigma)\mathbf{x}_0+\sigma\boldsymbol{\epsilon}\) with \(\boldsymbol{\epsilon}\sim\mathcal{N}(0,I)\), and the network regresses the velocity \(\boldsymbol{\epsilon}-\mathbf{x}_0\). Noise levels are sampled independently for video blocks and action tokens. We write \(\mathcal{L}^{e}_{\mathrm{vid}}\) and \(\mathcal{L}^{e}_{\mathrm{act}}\) for the flow-matching losses on the future head camera latents and on the action chunk of embodiment \(e\in\{\mathrm{r},\mathrm{h}\}\).

\textbf{Robot pre-training. }
We first train on robot demonstrations for 20,000 steps with just robot stream:
\[
\mathcal{L}_{\mathrm{pre}}
=
\mathcal{L}^{\mathrm{r}}_{\mathrm{vid}}
+
\lambda_a\mathcal{L}^{\mathrm{r}}_{\mathrm{act}}.
\]

\textbf{Joint co-training. }
We then fine-tune the model on aligned human-robot pairs. The two streams share model parameters but remain \emph{attention-isolated}: human tokens attend only within the human stream, and robot tokens only within the robot stream. Each stream predicts its own future video and actions:
\[
\mathcal{L}_{\mathrm{joint}}
=
\mathcal{L}^{\mathrm{r}}_{\mathrm{vid}}
+\lambda_a\mathcal{L}^{\mathrm{r}}_{\mathrm{act}}
+\lambda_{\mathrm{hv}}\mathcal{L}^{\mathrm{h}}_{\mathrm{vid}}
+\lambda_{\mathrm{ha}}\mathcal{L}^{\mathrm{h}}_{\mathrm{act}}.
\]
Thus, human demonstrations affect the robot pathway only through the shared model parameters while preserving the robot-only inference pathway.

\textbf{Human-to-robot style transfer. }
The second fine-tuning objective explicitly couples the aligned streams by predicting the robot future video from the corresponding human video clip. The full human clip is provided as clean context, while the robot future latents are noised and denoised:
\[
\mathcal{L}_{\mathrm{ST}}
=
\mathbb{E}_{\sigma,\boldsymbol{\epsilon}}
\left\|
v_\theta
\left(
\mathbf{x}^{\mathrm{r}}_\sigma
\,\middle|\,
\mathbf{x}^{\mathrm{h}}_0,\mathbf{c}^{\mathrm{r}}
\right)
-
(\boldsymbol{\epsilon}-\mathbf{x}^{\mathrm{r}}_0)
\right\|_2^2,
\]
where \(\mathbf{c}^{\mathrm{r}}\) denotes the robot CLIP context. To prevent the model from bypassing the human stream, we mask the robot conditioning latent and zero both action/state registers. The robot retains only the CLIP embeddings of its initial head and wrist observations for coarse scene context. This encourages the robot future prediction to rely on the aligned human trajectory together with the robot scene context. Only the robot video is supervised under this objective.

During fine-tuning, we train for 1,000 steps and alternate between \(\mathcal{L}_{\mathrm{joint}}\) and \(\mathcal{L}_{\mathrm{ST}}\), each with probability \(0.5\). The joint objective learns future motion and dynamics within each embodiment, while the style-transfer objective establishes cross-embodiment associations. At deployment, only the robot stream is used. The policy predicts 16-step action chunks and synchronously executes the first 8 actions before replanning. See Appendix \Cref{app:method} for more details on the training procedure.

%% file: text/eval.tex
\section{Experiments}
\label{sec:eval}

\begin{figure}[t]
  \centering
  \includegraphics[width=\textwidth]{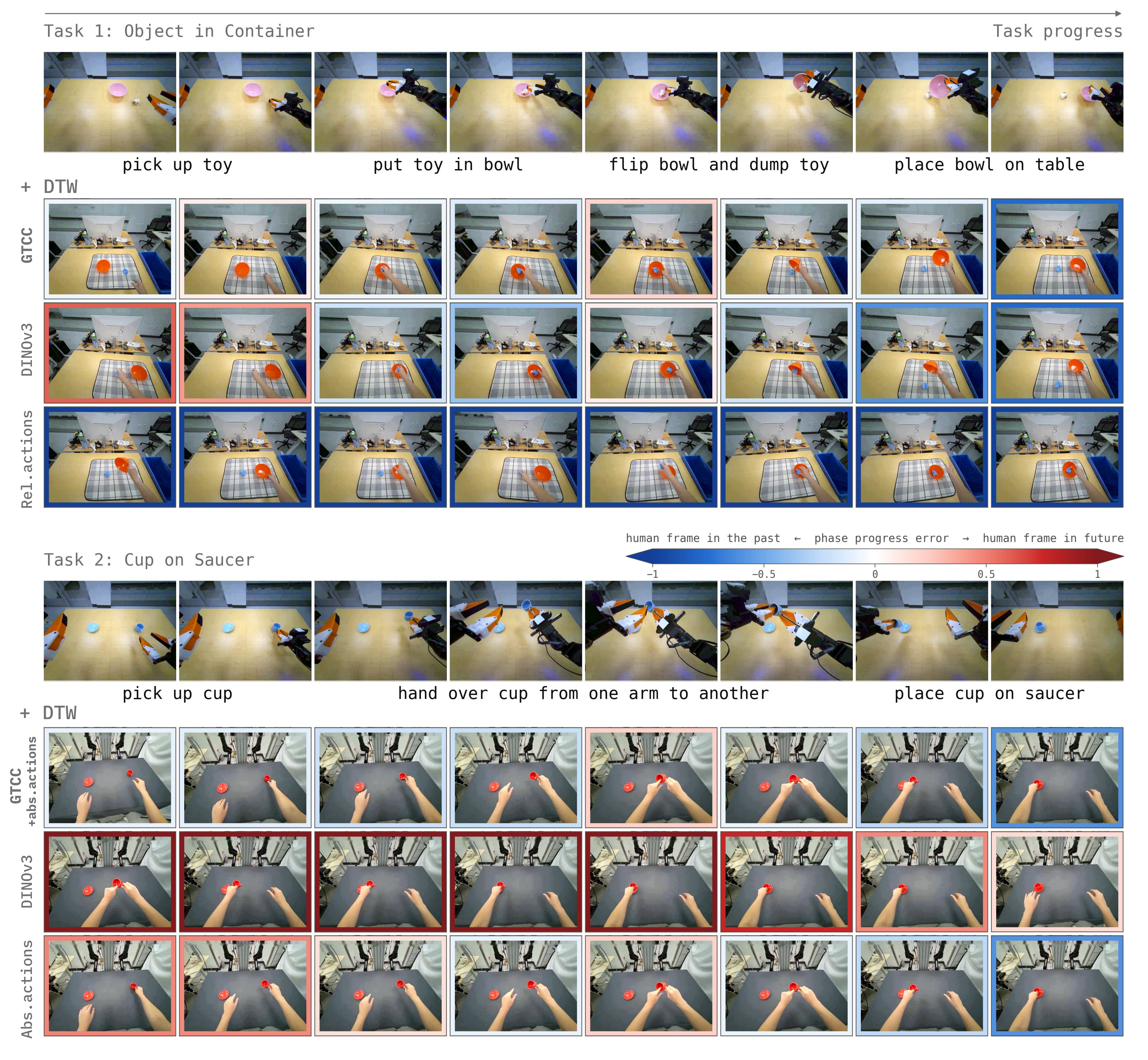}
  \vspace{-8mm}
  \caption{
  \textbf{Qualitative alignment results.}
    For each task, the top row shows robot query frames and the rows below show the corresponding human frames retrieved by each method. Border color denotes signed phase-progress error (blue: lagging, white: matched, red: ahead). Our method retrieves human frames that more closely follow the robot's task progress.}
  \label{fig:alignment}
  \vspace{-4mm}
\end{figure}


\subsection{Behavior Alignment Evaluation}
\label{sec:alignment-eval}

We evaluate whether the learned representation supports temporal alignment between robot and human demonstrations. Given a robot query segment, we retrieve and align the corresponding segment from human demonstrations, and evaluate the resulting frame-level correspondences against ground-truth phase annotations.

\textbf{Tasks and evaluation protocol. }
We evaluate on two tasks with distinct temporal structures and embodiment-dependent speed differences, shown in \Cref{fig:alignment}. In \emph{Object in Container}, each episode contains 6-7 repetitions of the four-phase manipulation cycle, creating multiple valid occurrences of the same phase within a trajectory. Human demonstrations execute the cycle approximately \(2\times\) faster than the robot. In \emph{Cup on Saucer}, each episode progresses through three phases, with the human operating \(3.4\times\) faster than the robot. For both tasks, we evaluate against an in-domain (ID) human set collected in the same environment as the robot demonstrations and an out-of-distribution (OOD) set collected in different scenes.
For each robot episode, we sample 400 evenly spaced query windows. Each query is aligned against every human episode in the candidate set. We retain the minimum-cost alignment from each human episode and average the resulting alignment metrics. Details of the evaluation dataset can be found in the Appendix \Cref{app:tasks} and \Cref{app:alignment-scenarios}.

We compare several representation and matching variants.
\underline{GTCC w/ DTW} applies DTW to temporal embeddings learned with GTCC, while \underline{GTCC w/o DTW} uses the same embeddings with rigid sliding, isolating the effect of temporal warping.
\underline{GTCC + rel./abs.\ actions} concatenates each GTCC feature with a 7-DoF end-effector pose per arm before applying DTW. We consider either absolute poses or poses expressed relative to the first frame of the query window.
\underline{Rel./abs.\ actions} performs DTW using these pose features alone, isolating the contribution of action information without GTCC features.
\underline{DINOv3 w/ DTW} uses frozen DINOv3 ViT-S/16~\citep{simeoni2025dinov3} class-token features to compare GTCC against a generic visual representation, and \underline{DINOv3 + rel./abs.\ actions} augments them with the corresponding end-effector poses described above.
\underline{Proportional mapping} matches frames according to normalized temporal position within the target episode.

\textbf{Phase overlap.}
We measure coarse phase alignment as the fraction of matched frame pairs that share the same phase label, i.e., the number of correctly matched frames divided by the total number of matched frames.

\textbf{Phase progress error.}
Phase overlap does not distinguish between frames at different points within the same phase. To measure finer temporal alignment, we assign each frame a progress value that encodes both its phase and its progress within that phase. Specifically, for a frame $f$ in phase $k_m$, spanning frames $s_m,\dots,e_m$, we define
\begin{equation}
    p(f)=\frac{1}{K}
    \left(
        k_m+\frac{f-s_m}{e_m-s_m+1}
    \right),
    \qquad
    E=\frac{1}{|\mathcal{P}|}
    \sum_{(f_q,f_t)\in\mathcal{P}}
    \delta\!\left(p(f_q),p(f_t)\right).
\end{equation}
Here, $p(f)\in[0,1)$ represents the frame's normalized position within one task cycle, and $E$ is the average absolute progress difference between matched query and target frames. Thus, lower error indicates more precise temporal correspondence. We provide quantitative and qualitative results in \Cref{tab:alignment-results} and \Cref{fig:alignment}, respectively. More results can be found in Appendix \Cref{fig:more-alignment-task1} and \Cref{fig:more-alignment-task2}.

\definecolor{best}{RGB}{204,235,214}
\definecolor{second}{RGB}{232,245,236}
\definecolor{groupbg}{RGB}{245,245,245}

\newcommand{\bestcell}[1]{\cellcolor{best}#1}
\newcommand{\secondcell}[1]{\cellcolor{second}#1}

\begin{table*}[t]
\centering
\footnotesize
\renewcommand{\arraystretch}{1.05}

\begin{tabular*}{\textwidth}{@{\extracolsep{\fill}}lcccccccc@{}}
\toprule
 & \multicolumn{4}{c}{\textbf{Object in Container}}
 & \multicolumn{4}{c}{\textbf{Cup on Saucer}} \\
\cmidrule(lr){2-5}\cmidrule(lr){6-9}

 & \multicolumn{2}{c}{Overlap (\%) $\uparrow$}
 & \multicolumn{2}{c}{Prog.\ err.\ $\downarrow$}
 & \multicolumn{2}{c}{Overlap (\%) $\uparrow$}
 & \multicolumn{2}{c}{Prog.\ err.\ $\downarrow$} \\
\cmidrule(lr){2-3}\cmidrule(lr){4-5}
\cmidrule(lr){6-7}\cmidrule(lr){8-9}

\textbf{Method}
& \textbf{ID} & \textbf{OOD}
& \textbf{ID} & \textbf{OOD}
& \textbf{ID} & \textbf{OOD}
& \textbf{ID} & \textbf{OOD} \\
\midrule

\rowcolor{groupbg}
\multicolumn{9}{l}{\textit{GTCC variants}} \\

GTCC + DTW
& \cellcolor{best}\textbf{69.0}
& \cellcolor{best}\textbf{57.8}
& \cellcolor{second}\underline{.061}
& \cellcolor{best}\textbf{.098}
& 64.8
& \cellcolor{second}\underline{71.7}
& \cellcolor{second}\underline{.088}
& \cellcolor{second}\underline{.075} \\

\quad $(-)$ DTW
& 61.8 & 24.7 & .080 & .231
& 61.9 & 70.8 & .092 & .085 \\

\quad $(+)$ rel.\ actions
& \cellcolor{second}\underline{68.9}
& \cellcolor{second}\underline{54.9}
& \cellcolor{best}\textbf{.060}
& \cellcolor{second}\underline{.104}
& 61.8 & 62.7 & .100 & .123 \\

\quad $(+)$ abs.\ actions
& 43.7 & 54.6 & .124 & .106
& \cellcolor{best}\textbf{79.2}
& \cellcolor{best}\textbf{76.5}
& \cellcolor{best}\textbf{.070}
& \cellcolor{best}\textbf{.073} \\

\addlinespace[2pt]
\rowcolor{groupbg}
\multicolumn{9}{l}{\textit{Feature and action baselines}} \\

DINOv3 + DTW
& 39.4 & 42.3 & .148 & .156
& 48.9 & 22.1 & .294 & .433 \\

\quad $(+)$ rel.\ actions
& 35.1 & 43.9 & .171 & .153
& 49.1 & 25.3 & .282 & .377 \\

\quad $(+)$ abs.\ actions
& 22.1 & 35.6 & .234 & .178
& 45.7 & 33.8 & .222 & .317 \\

Rel.\ actions + DTW
& 25.6 & 39.6 & .229 & .169
& 34.2 & 29.2 & .390 & .433 \\

Abs.\ actions + DTW
& 17.1 & 33.8 & .264 & .183
& \cellcolor{second}\underline{70.4}
& 67.1 & .118 & .123 \\

\addlinespace[2pt]
\rowcolor{groupbg}
\multicolumn{9}{l}{\textit{Temporal baseline}} \\

Proportional mapping
& 34.9 & 25.5 & .194 & .218
& 32.5 & 60.4 & .217 & .139 \\

\bottomrule
\end{tabular*}

\caption{
\textbf{Cross-embodiment alignment results. }
We report frame-level overlap and phase-progress error on in-domain (ID) and out-of-domain (OOD) human demonstrations. GTCC yields the strongest visual-only alignment across both tasks, substantially outperforming DINOv3 features, while DTW further improves alignment, particularly for \textit{Object in Container}. Pose cues provide limited benefit when the action gap is large, but become more informative for \textit{Cup on Saucer}, where combining absolute end-effector poses with GTCC achieves the best overall performance.}
\vspace{-4mm}
\label{tab:alignment-results}
\end{table*}

\looseness-1
\textbf{GTCC provides substantially stronger cross-embodiment visual correspondence than generic visual features.}
Using visual features alone, \underline{GTCC + DTW} consistently outperforms \underline{DINOv3 + DTW} across both tasks and both splits. On \emph{Object in Container}, GTCC substantially improves both overlap and progress error over DINOv3 across ID and OOD settings. The gap is bigger on \emph{Cup on Saucer}, where DINOv3's overlap drops sharply under OOD scene changes, while GTCC remains robust at 71.7\% OOD overlap. These results suggest that temporal cycle consistency learns representations that transfer across embodiment and scene changes more reliably than generic visual features.

\looseness-1
\textbf{Temporal warping improves alignment under execution-speed differences.}
On \emph{Object in Container}, removing DTW substantially degrades performance, with OOD overlap dropping from 57.8\% to 24.7\% and progress error increasing from .098 to .231. DTW also provides smaller but consistent gains in the remaining settings, showing that non-uniform temporal warping generally helps align human and robot trajectories that execute the same behavior at different rates.

\looseness-1
\textbf{Action information helps only when the gap is small across embodiments.}
On \emph{Object in Container}, adding pose features provides little benefit over GTCC alone, and pose-only baselines are substantially weaker. In contrast, on \emph{Cup on Saucer}, absolute poses are highly informative, and combining them with GTCC yields the best performance. This suggests that pose cues are most useful when human and robot trajectories share a meaningful spatial correspondence, while GTCC remains effective regardless.

\subsection{Policy Learning}

\begin{figure}[t]
  \centering
  \includegraphics[width=\textwidth]{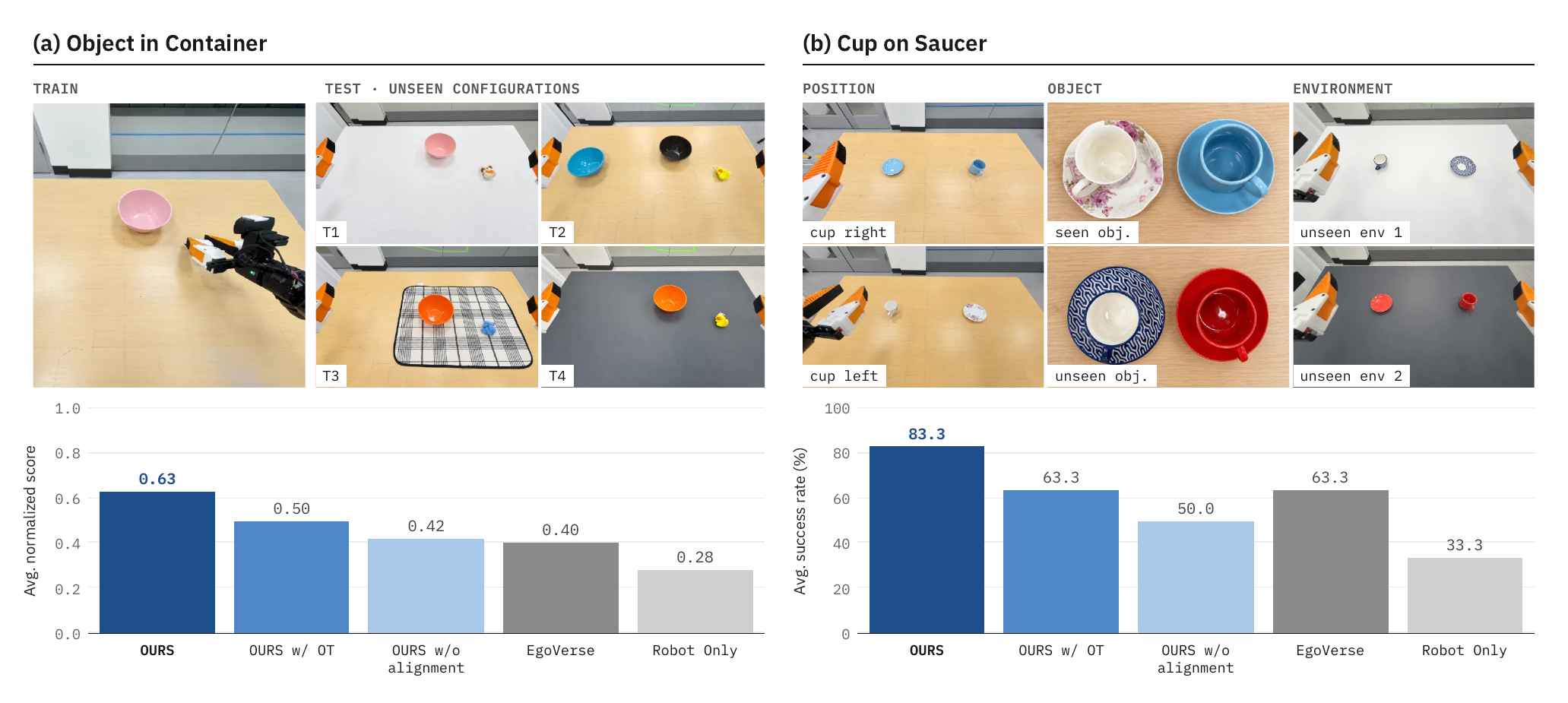}
  \vspace{-8mm}
  \caption{
  \textbf{Real-world rollout results.}
    We evaluate \emph{Object in Container} across four unseen configurations including unseen background, target objects, and distractor objects, and \emph{Cup on Saucer} across unseen objects and environments. Our method achieves the strongest overall performance.}
    \vspace{-6mm}
  \label{fig:policy}
\end{figure}

To evaluate the effectiveness of behavior-aligned human-robot demonstrations and our cross-embodiment training objectives, we test policy on two real-world manipulation tasks. For \textit{Object in Container}, which is cyclic, we use a fine-grained score to capture partial task progress. The policy continuously executes the four-skill sequence shown in \Cref{fig:alignment} for 60 seconds, receiving one point for each successfully completed skill. We evaluate four OOD settings, shown in \Cref{fig:policy} (a), with 20 rollouts in total, randomizing the initial object positions in each rollout. Following EgoVerse \citep{punamiya2026egoverse}, we report the score normalized by the average in-domain performance (e.g., 55 points). For \textit{Cup on Saucer}, we evaluate 30 rollouts across OOD scenarios with unseen objects and environments, as shown in \Cref{fig:policy} (b), and report the number of successful trials. We compare against the following baselines and ablations:

\underline{OURS w/o alignment.} We remove behavior alignment and instead randomly sample human and robot trajectories from their respective datasets.

\underline{OURS w/ OT.} We replace the human-to-robot style transfer objective with an optimal transport (OT) loss applied to the input observation tokens, including image and state tokens, across the human and robot streams, following a formulation similar to EgoBridge~\citep{punamiya2025egobridge}.

\underline{EgoVerse~\citep{punamiya2026egoverse}.} We remove the cross-embodiment style-transfer objective and simply co-train on human and robot data using the action and video prediction objectives.

\underline{Robot only.} We use the pretrained robot policy trained solely on in-domain robot demonstrations, without finetuning with human data.

Our results in \Cref{fig:policy} highlight the complementary benefits of human data finetuning, behavior alignment, and explicit cross-embodiment transfer objective. \underline{EgoVerse} improves over \underline{Robot Only} on both tasks, showing that co-training with human demonstrations is beneficial. However, \underline{OURS} achieves the strongest performance on both tasks, indicating that explicitly coupling the human and robot streams through the style-transfer objective provides additional gains beyond co-training alone.

Replacing the style-transfer objective with an OT loss in \underline{OURS w/ OT} also yields strong performance, outperforming \underline{EgoVerse} on \textit{Object in Container} and matching its performance on \textit{Cup on Saucer}, while still falling short of \underline{OURS}. This suggests that our style-transfer objective provides a more effective cross-embodiment learning signal than directly matching input representations with an OT loss. Meanwhile, removing behavior alignment in \underline{OURS w/o alignment} degrades performance on both tasks. This underscores the importance of pairing human and robot trajectories based on behavioral correspondence when applying cross-embodiment supervision. Overall, \textbf{behavior alignment and the cross-embodiment style-transfer objective provide complementary gains for human-to-robot transfer.}

%% file: text/addn.tex
\vspace{-1mm}
\section*{Acknowledgments}
\vspace{-2mm}
The authors would like to thank Akshara Rai, Jimmy Yang, and Homanga Bharadhwaj for insightful brainstorming and ideation; Josh Citron for hardware assistance; Chen Chen and all REAL Lab members for valuable technical discussions and feedback on the paper. This work was supported in part by the NSF Award \#2143601, \#2037101, and \#2132519, Apple, We would like to thank Google and TRI for the UR5 robot hardware, ARX for the ARX robot hardware. The views and conclusions contained herein are those of the authors and should not be interpreted as necessarily representing the official policies, either expressed or implied, of the sponsors.

\vspace{-1mm}
\section{AI use statement}
\vspace{-2mm}
Claude Design Opus 5.5 was used to assist with layout design, visual element generation, and image editing (e.g., brightening) for \Cref{fig:method}, \Cref{fig:alignment}, \Cref{fig:policy}, and \Cref{fig:alignment-scenarios}. ChatGPT was used to identify relevant prior works and references and to refine wording and sentence structure. Claude Code Opus 5.5 was used to assist with data visualization and data processing scripts, navigating and understanding the DreamZero~\citep{ye2026world} codebase, and managing evaluation experiments reported in \Cref{tab:alignment-results}. All technical decisions, experiments, analyses, and final manuscript content were reviewed by the authors and validated.

\vspace{-1mm}
\section{Ethics statement}
\vspace{-2mm}
This work studies cross-embodiment robot learning from human and robot demonstrations. The main ethical considerations concern human data privacy and robot safety. Human demonstrations, particularly egocentric recordings, may inadvertently capture bystanders or sensitive information, motivating appropriate consent, careful data handling, and review before data release. In addition, learned robot policies may behave unpredictably outside the training distribution, so real-world deployment should include appropriate human supervision. More broadly, while learning from human data may reduce the cost of robot data collection and enable more capable robotic systems, these capabilities could also be applied in settings that raise privacy, safety, labor, or surveillance concerns, making responsible data collection, evaluation, and deployment important considerations for future work.

\vspace{-1mm}
\section{Reproducibility statement}
\vspace{-2mm}
We have made extensive efforts to ensure the reproducibility of our work. Details of the model architecture, training objectives, and hyperparameters are provided in \Cref{sec:method} in the main paper and in Appendix \Cref{app:method}. The datasets used in the real-world experiments are described in Appendix \Cref{app:setup} and \Cref{app:data}. Evaluation metrics and baseline implementations are specified in \Cref{sec:eval} in the main paper. We also provide an anonymous project website (\url{https://batch-embodiment.github.io/}) with qualitative results and visualizations, and we will release code, datasets, and model checkpoints upon acceptance to further support reproducibility.

%% file: text/supp.tex
\begin{center}
    {\Large\bfseries Appendix}
\end{center}
\label{sec:supp}


\providecommand{\placeholder}[1]{\textcolor{red}{#1}}

\section{Data Collection Details}
\label{app:setup}

\begin{wrapfigure}{r}{0.50\textwidth}
    \centering
    \vspace{-6mm}
    \includegraphics[width=0.48\textwidth]{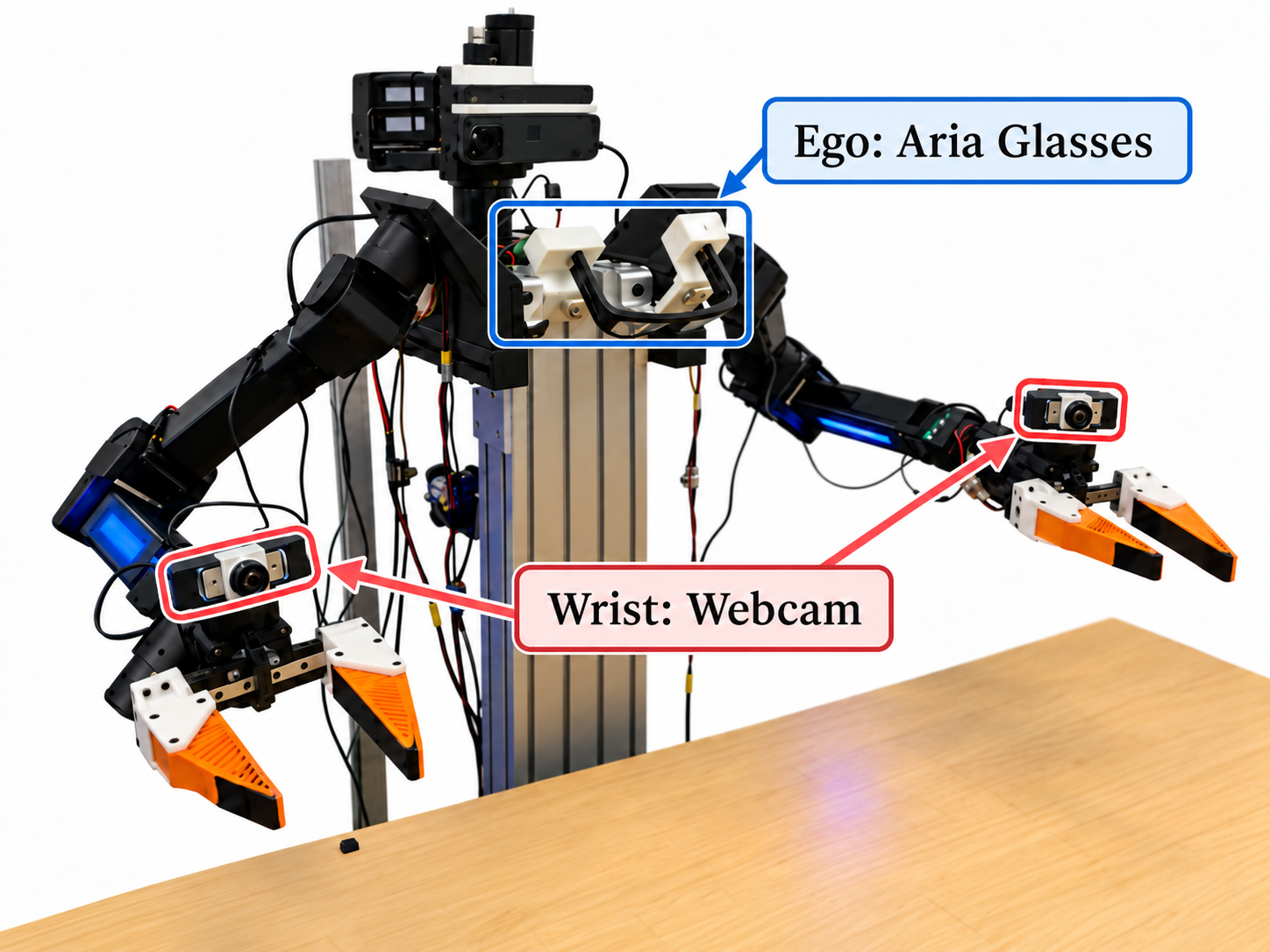}
    \caption{
        \textbf{Robot hardware setup.}
    }
    \label{fig:robot-hardware}
    \vspace{-4mm}
\end{wrapfigure}

Human demonstrations are recorded using the main RGB camera of Aria Gen~1 glasses worn by the demonstrators. As shown in \Cref{fig:robot-hardware}, our robot platform consists of two ARX5 arms side-mounted on a custom 3D-printed shoulder structure, providing a human-like bimanual workspace. To obtain corresponding robot observations, we mount an Aria Gen~1 glass on the robot for the head-camera view and a Logitech RGB webcam on each wrist for close-up manipulation views.

\looseness-1
\textbf{Human hand poses. }
Human hand poses are obtained using the Machine Perception Services (MPS) provided by Aria Gen 1. The hand poses are transformed into the coordinate frame of the tracked head camera. Each hand is represented by a 3D position and a quaternion orientation, resulting in a 7-dimensional representation per hand. Because no gripper-width equivalent is available in the human data, we zero-pad this dimension and exclude it from the human action loss during training.

\textbf{Robot proprioception and actions. }
Robot actions are derived from proprioceptive measurements of both arms. For each arm, we use the 3D end-effector position, quaternion orientation, and gripper width, resulting in an 8-dimensional representation per arm and a 16-dimensional bimanual action representation. The end-effector poses are transformed into the robot head-camera coordinate frame using calibrated extrinsics.

\section{Method Details}
\label{app:method}
\vspace{-1mm}

We evaluate on two tasks: single-arm \textit{Object in Container}
and bimanual \textit{Cup on Saucer}. The model, objectives, and training recipe are identical; what changes is the action dimension, the number of wrist camera inputs, and hence the width of the conditioning canvas and the
token sequence length. The table below summarizes these differences.

\begin{table}[h]
  \centering
  \small
  \begin{tabular}{@{}lll@{}}
    \toprule
     & \textbf{Object in Container} & \textbf{Cup on Saucer} \\
    \midrule
    Embodiment              & single arm & bimanual \\
    State / action dim      & 8 (pos 3, quat 4, gripper 1) & 16 ($2\times8$) \\
    Human action dim        & 7 (one hand) & 14 (two hands) \\
    Wrist cameras           & 1 & 2 \\
    Conditioning canvas     & $240\times560$ & $240\times800$ \\
    Clean tokens / frame    & 525 & 750 \\
    Noisy tokens / frame    & 300 & 300 \\
    Robot CLIP tokens       & 514 ($2\times257$) & 771 ($3\times257$) \\
    Human CLIP tokens       & 257 & 257 \\
    Joint co-training pass  & 2{,}284 tokens & 2{,}734 tokens \\
    Style-transfer pass     & 3{,}334 tokens & 4{,}234 tokens \\
    Deployment              & 1{,}142 tokens & 1{,}367 tokens \\
    \midrule
    robot pretrain demos  & 90 (28{,}740 windows) & 270 (25{,}480 windows) \\
    human fine-tune demos  & 5 per scenario & 8 per scenario \\
    Robot demo duration   & $\sim$12 s & $\sim$15 s \\
    Human demo duration   & $\sim$7 s & $\sim$4 s \\
    \bottomrule
  \end{tabular}
  \vspace{1mm}
  \caption{Per-task configuration.}
  \label{app:tasks}
\end{table}

\subsection{Behavior Alignment Details}
\label{app:behavior-alignment}

Human-robot pairs are computed once per task before policy training.
Every frame of the robot and human recordings is embedded with the
GTCC encoder described in \Cref{subsec:behavior-alignment}. Given a robot query window
$u_{1:L}$ and a candidate human sequence $v_{1:T}$, we define the
pairwise embedding cost as
\[
C_{ij} = \|u_i - v_j\|_2.
\]

We apply subsequence dynamic time warping (DTW) to find the
minimum-cost human subsequence that aligns to the full robot query
while allowing non-uniform temporal warping. Let $D[i,j]$ denote the
cumulative DTW cost. We select the endpoint
\[
j^\star = \arg\min_j D[L,j],
\]
and backtrack from $(L,j^\star)$ to recover the matched human
subsequence and frame-level warping path.

For each robot window, we rank human candidates by normalized DTW cost
and store the top-$k$ matches together with their warping paths in raw
frame units ($k=5$ for \textit{Object in Container} and $k=1$ for
\textit{Cup on Saucer}). At each training step, one candidate is sampled
uniformly, and the corresponding human frames, states, and actions are
read at the indices specified by the warping path, effectively
time-warping the human trajectory onto the robot window.

\textbf{Computational cost. }
Let $N$ and $M$ denote the numbers of robot and human episodes, respectively, and $T$ the episode length in frames. Each GTCC training epoch pairs every episode in the larger dataset with one from the other embodiment, yielding a cost of $\mathcal{O}(\max(N,M)\,T^2)$, linear in the larger dataset size.

Retrieval costs $\mathcal{O}((N+M)\,T + NMLT^2)$: embedding all episodes once takes $\mathcal{O}((N+M)\,T)$, while matching $\mathcal{O}(NT)$ robot query windows of length $L$ against $M$ human episodes via subsequence DTW takes $\mathcal{O}(NMLT^2)$. These matches are independent and readily parallelizable.

For $N=M=30$ and $T\approx200$, training takes about 100 s per epoch on one GPU, or 5.5 GPU-hours for 200 epochs. On the scale of our datasets shown in \Cref{app:tasks}, retrieval takes a few minutes for both embedding and subsequence DTW matching on a single CPU core.

\subsection{Cross-embodiment Video Action Model}
\label{app:video-action-model}

\subsubsection{Data Processing}
\label{app:data}

\paragraph{Temporal sampling.}
Recordings are at 30\,fps. A training window takes 9 head-camera frames at
a stride of 6 (5\,fps, ${\approx}1.6$\,s) and 16 actions at a stride of 3
(10\,Hz), spanning the same interval.

\paragraph{Image geometry.}
Head-camera frames are resized to $240\times320$. Each wrist image ($224\times224$) is zero-padded to $240\times240$ and concatenated with the head view along their latent width dimension. This produces an equivalent $240\times560$ canvas for \textit{Object in Container}, which uses one wrist view, and a $240\times800$ canvas for \textit{Cup on Saucer}, which uses two wrist views. Human streams contain only the head view, and the corresponding wrist regions are zero-padded. After $8\times$ spatial VAE compression and $2\times2$ patchification, a head-only noisy frame produces $15\times20=300$ tokens, while the clean conditioning frame produces 525 and 750 tokens for \textit{Object in Container} and \textit{Cup on Saucer}, respectively. Following Wan2.1-I2V \citep{wan2025wan}, each video token contains 36 input channels: 16 noisy latent channels, a 4-channel mask indicating the known conditioning frame, and the 16-channel VAE latent of that frame.

\paragraph{States and actions.}
Robot states and actions use 8 dimensions per arm (position, orientation quaternion, gripper width): 8 dimensions for \textit{Object in Container} and 16 for \textit{Cup on Saucer}. Human hand poses use the same per-arm layout without the gripper dimension, which is zero-padded and masked out of the human action loss. Each arm's pose trajectory is expressed relative to that arm's pose at the first frame, $\tilde{T}_t = T_0^{-1} T_t$, while the gripper stays absolute. All dimensions are normalized with 1st/99th-percentile statistics computed per embodiment.

\paragraph{Augmentation.}
Training video receives random crops (scale 0.95), color jitter
(brightness 0.3, contrast 0.4, saturation 0.5, hue 0.08),
random grayscale ($p = 0.1$), and random posterization to 4 bits
($p = 0.1$). One set of parameters is sampled per clip and shared by all
frames and by the head and wrist views. Deployment inputs are not augmented.

\subsubsection{Attention Structure}
\label{app:attention}

Each embodiment is represented as a separate stream consisting of clean conditioning tokens, noisy future-video tokens, state tokens, and action tokens. Within each stream, we use block-causal self-attention over the conditioning context and the predicted block. Noisy video tokens attend to the stream's clean conditioning context and to the noisy video tokens within the same block. Action tokens additionally attend to the corresponding noisy video block, state token, and action tokens.

Positional information is also kept stream-specific: rotary position indices for both image tokens and register tokens (e.g., state, action tokens) restart from zero for each embodiment. Cross-attention to the CLIP image features of the conditioning frame is likewise separated by embodiment, such that each stream attends only to its own conditioning image features.

The two fine-tune objectives differ only in how clean-context attention is shared across the human and robot streams:

\begin{itemize}[leftmargin=*]
\item \textbf{Joint co-training.}
Human and robot streams are attention-isolated. Each stream attends only to its own clean conditioning context, with separate rotary positions and CLIP conditioning. As a result, the robot predictions are independent of the presence of the human stream, while both streams are optimized jointly within the same model.

\item \textbf{Style transfer.}
The human stream is placed before the robot stream, and the robot stream attends to the clean human video context together with its robot CLIP conditioning; the robot conditioning latent and registers are masked.
\end{itemize}

\subsubsection{Training Procedure}
\label{app:training}

\textbf{Flow matching.}
We train video and action prediction with flow matching. Given a clean target $x_0$, we construct
\[
x_\sigma = (1-\sigma)x_0 + \sigma\epsilon,
\qquad
\epsilon \sim \mathcal{N}(0,I),
\]
and train the network to predict the velocity $\epsilon-x_0$.
We denote the corresponding video and action losses for embodiment
$e\in\{r,h\}$ as $\mathcal{L}^{e}_{\mathrm{vid}}$ and
$\mathcal{L}^{e}_{\mathrm{act}}$.

Noise levels follow a flow-matching schedule with shift 5 over 1,000
discrete timesteps. Video timesteps are drawn as
$t=\lfloor(1-u)\cdot1000\rfloor$ with $u\sim\mathrm{Beta}(\alpha,1)$
and are shared across frames within a causal block. Action timesteps are
drawn uniformly and independently for each action token. The conditioning
latent frame remains clean and is excluded from the video loss, which is
computed only on future head-camera latents.

\textbf{Robot pretraining.}
We first train a single robot stream for 20,000 steps with
\[
\mathcal{L}_{\mathrm{pre}}
=
\mathcal{L}^{r}_{\mathrm{vid}}
+
\lambda_a \mathcal{L}^{r}_{\mathrm{act}}.
\]
Pretraining initializes from the first eight blocks of Wan2.1-I2V-14B
and jointly trains all transformer blocks, the patch embedding, timestep
and image-embedding projections, output head, and action/state encoders
and decoder. Training uses 4 NVIDIA RTX PRO 6000 Blackwell Server Edition
96GB GPUs, a batch size of 16 per GPU, a learning rate of $1\times10^{-4}$,
and video timesteps sampled from $\mathrm{Beta}(3,1)$.

\textbf{Cross-embodiment fine-tuning.}
Fine-tuning initializes from the pretrained checkpoint and runs for 1,000
steps on aligned human-robot pairs. For joint co-training, the two streams
remain attention-isolated and optimize
\[
\mathcal{L}_{\mathrm{joint}}
=
\mathcal{L}^{r}_{\mathrm{vid}}
+\lambda_a\mathcal{L}^{r}_{\mathrm{act}}
+\lambda_{hv}\mathcal{L}^{h}_{\mathrm{vid}}
+\lambda_{ha}\mathcal{L}^{h}_{\mathrm{act}}.
\]
Human video and action timesteps are sampled independently from the same
distributions as the robot stream.

For human-to-robot style transfer, the human latent frames are provided
as clean context, while the robot future latents are noised and denoised.
The robot conditioning latent and action/state registers are masked, and
only the robot video prediction is supervised. During fine-tuning, each
step optimizes either $\mathcal{L}_{\mathrm{joint}}$ or
$\mathcal{L}_{\mathrm{ST}}$ with probability 0.5.

Fine-tuning uses a learning rate of $2\times10^{-5}$ with layer-wise decay
0.95. Video timesteps are sampled from $\mathrm{Beta}(1.5,1)$. In style
transfer, all human latent frames are fixed at $t=0$, while robot latent
frames are noised according to the video prior.

\subsubsection{Inference}
\label{app:inference}

At deployment the sequence contains only the robot stream: the current head and wrist images are encoded once into a key-value cache as the clean conditioning frame, and the policy denoises one video block together with the 16 action tokens, conditioned on the current state.

We sample with a 16-step UniPC multistep flow solver (shift 5). The transformer is evaluated on 8 of the 16 steps under a fixed schedule and the remaining steps reuse the most recent velocity prediction. The action schedule runs to
$\sigma = 0$, while the video schedule is rescaled to end at $\sigma = 0.25$. We execute the first 8 of the 16 predicted actions at 10\,Hz, then reset the cache and replan from a new observation. The inference latency is 0.35s for single arm policy and 0.42s for bimanual policy on one GeForce RTX 5090 GPU.



\section{Additional Alignment Results}
\vspace{-1mm}
\label{app:alignment-results}

\subsection{Evaluation Metrics}
\label{app:alignment-metrics}

\textbf{Phase overlap.}
We measure coarse temporal alignment as the fraction of matched
frame pairs that share the same ground-truth phase label.

\textbf{Phase progress error.}
To capture finer temporal correspondence within a phase, we assign
each frame $f$ in phase $k_m$, spanning frames $s_m,\ldots,e_m$, a
normalized task-progress value
\[
p(f)
=
\frac{1}{K}
\left(
k_m + \frac{f-s_m}{e_m-s_m+1}
\right),
\]
where $K$ is the number of phases in one task cycle. For the set of
aligned frame pairs $\mathcal{P}$, we compute
\[
E
=
\frac{1}{|\mathcal{P}|}
\sum_{(f_q,f_t)\in\mathcal{P}}
\delta\!\left(p(f_q),p(f_t)\right),
\]
where $\delta$ denotes the absolute progress difference. Lower error
indicates more precise temporal correspondence.

\subsection{Evaluation Scenarios}
\label{app:alignment-scenarios}

\begin{figure*}[t]
    \centering
    \includegraphics[width=\textwidth]{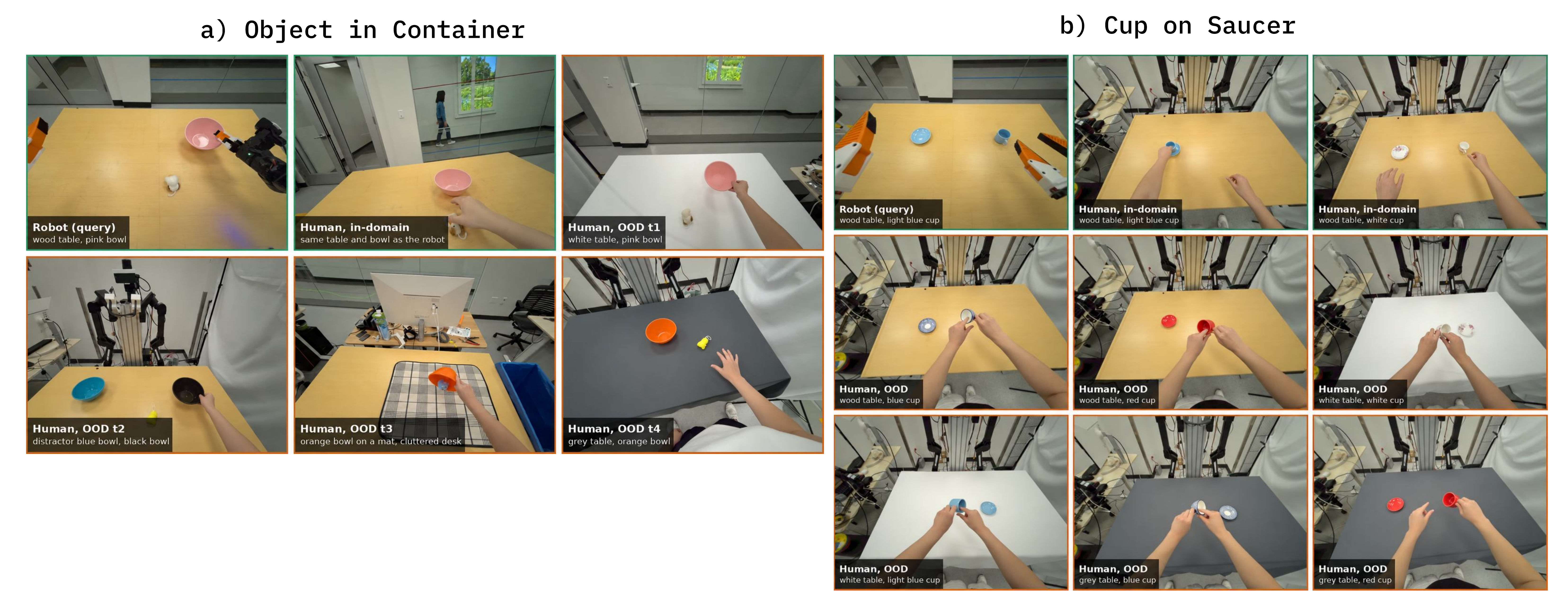}
    \vspace{-4mm}
    \caption{
        Alignment evaluation scenarios for \textit{Object in Container} and \textit{Cup on Saucer}.
    }
    \label{fig:alignment-scenarios}
\end{figure*}

We evaluate GTCC alignment using robot query trajectories and human
demonstrations collected under both in-domain (ID) and out-of-domain (OOD)
conditions. \Cref{fig:alignment-scenarios} visualizes the evaluation scenarios for the two tasks. For \textit{Object in Container}, the ID human demonstrations use the same
table and objects as the robot demonstrations despite viewpoint differences, while the OOD sets introduce
additional changes in object appearance, table color, and background. For
\textit{Cup on Saucer}, the ID demonstrations are collected on the same
wooden table as the robot data using the in-domain light-blue and white
cup-saucer sets. The OOD evaluation varies both object and scene appearance,
including unseen cup-saucer sets (e.g., red and blue) and white or gray
table cloths. For each scenario, we use 10 robot demonstrations and all available human
demonstrations for each scenario to compute the alignment evaluation metrics.

\subsection{Qualitative Alignment Results}
\label{app:alignment-qualitative}

\begin{figure*}[h]
    \centering
    \includegraphics[width=\textwidth]{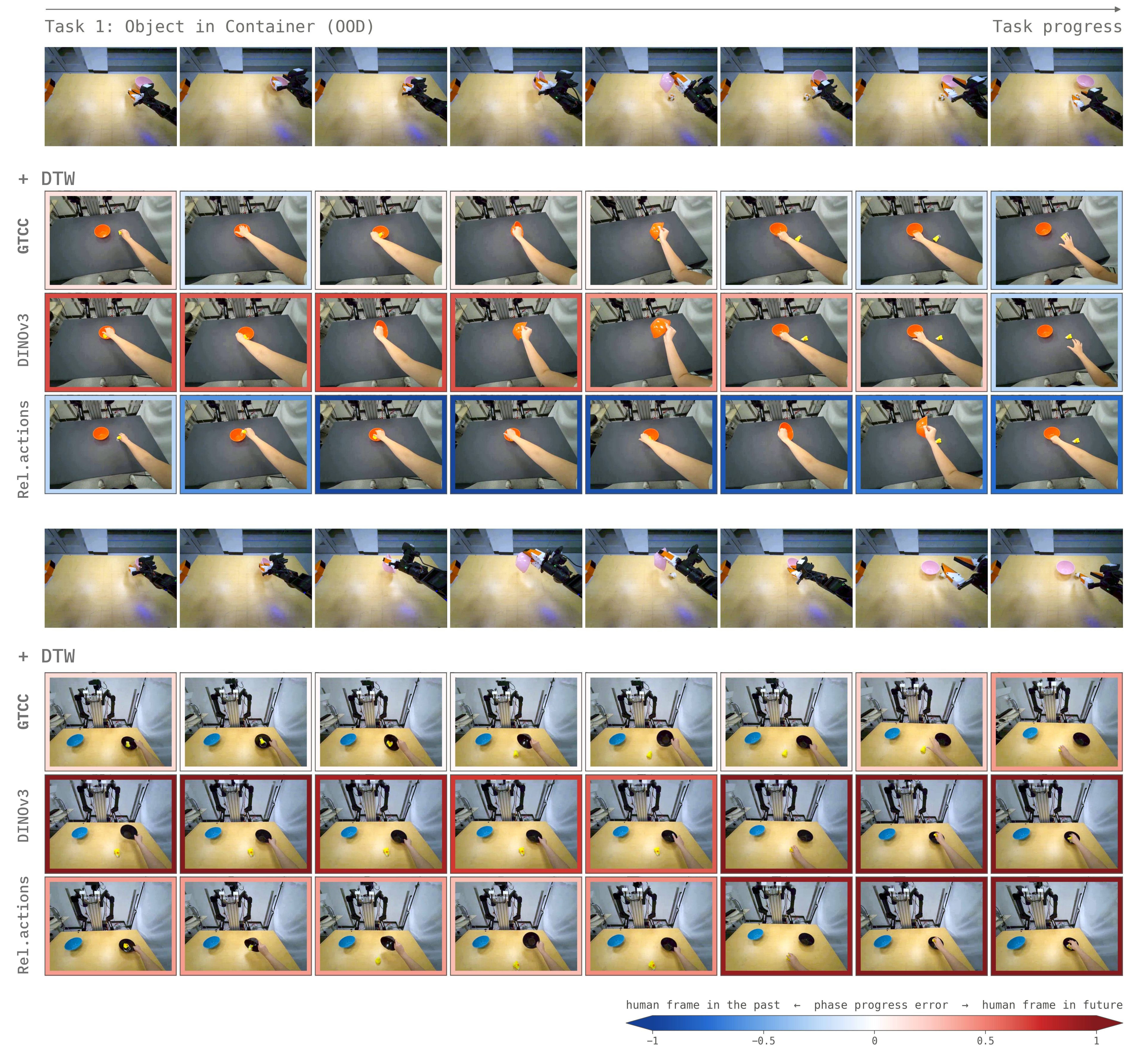}
    \caption{
        Additional qualitative alignment examples for \textit{Object in Container}.
    }
    \label{fig:more-alignment-task1}
\end{figure*}

\begin{figure*}[h]
    \centering
    \includegraphics[width=\textwidth]{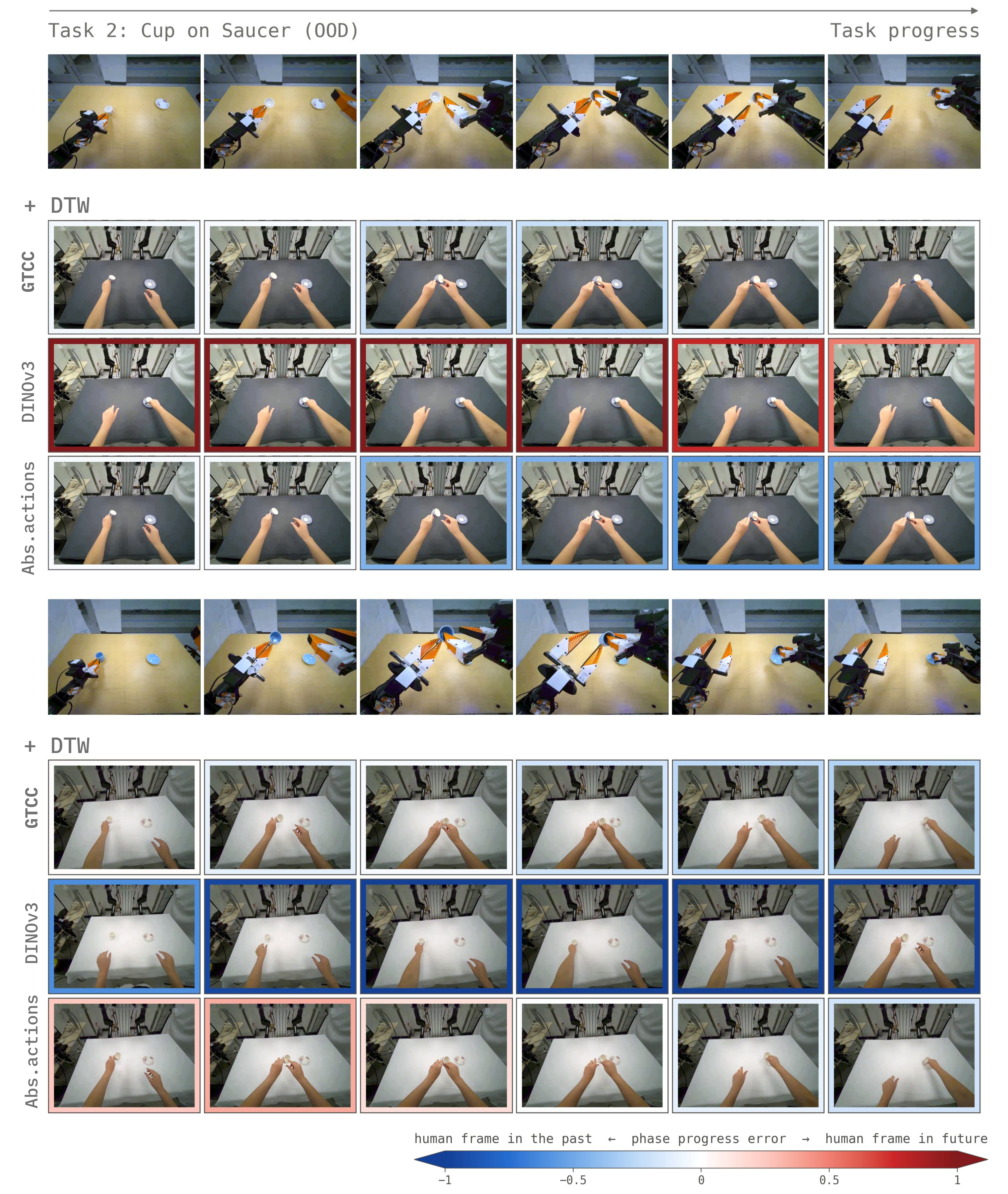}
    \caption{
        Additional qualitative alignment examples for \textit{Cup on Saucer}.
    }
    \label{fig:more-alignment-task2}
\end{figure*}

\Cref{fig:more-alignment-task1,fig:more-alignment-task2} show additional qualitative alignment examples for both tasks. GTCC consistently retrieves human frames that are more closely aligned in behavioral progress with the robot query.

\begin{figure}[t]
    \centering
    \includegraphics[width=\textwidth]{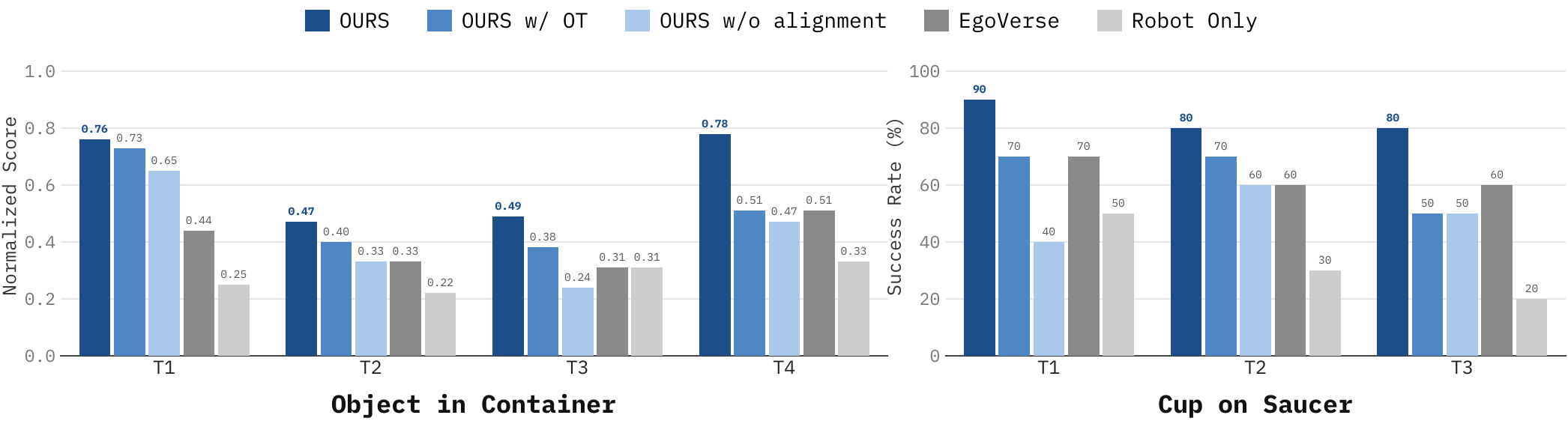}
    \caption{Policy performance per scenario. }
    \label{fig:scenario-results}
\end{figure}

\section{Additional Policy Results}
\label{app:policy-results}

\vspace{-1mm}
\subsection{Per-Scenario Performance}

We conduct 5 one-minute rollouts per scenario for \textit{Object in Container} and 10 rollouts per scenario for \textit{Cup on Saucer}. For \textit{Cup on Saucer},
\textsc{T1} tests object generalization using only unseen object sets
(red and blue), balanced across left- and right-side cup placements.
\textsc{T2} introduces a white tablecloth and mixes unseen object sets
(e.g., red and blue) with seen object sets (e.g., light blue and white), while evenly varying
the cup side across rollouts.
\textsc{T3} uses the same four object sets on a gray tablecloth, with
5 rollouts using seen objects and 5 using unseen objects, and an even split between left- and right-side cup placements.

\Cref{fig:scenario-results} breaks down policy performance across
individual evaluation scenarios. Across all scenarios on both tasks, \underline{OURS} achieves the strongest performance, with especially large gains over \underline{EgoVerse} under more background distribution shifts (e.g., T1, T4 in \textit{Object in Container} and T2, T3 in \textit{Cup on Saucer}). The consistent drop for \underline{OURS w/o alignment} further shows that behavior alignment is important for effectively transferring information from human demonstrations.

\clearpage